\documentclass[runningheads]{llncs}
\usepackage{graphicx}
\usepackage{amsmath,amssymb} 
\usepackage{color}
\usepackage[ruled,vlined]{algorithm2e}
\usepackage[width=122mm,left=12mm,paperwidth=146mm,height=193mm,top=12mm,paperheight=217mm]{geometry}
\usepackage{multirow}
\usepackage{array}
\usepackage{epsfig}
\usepackage{epstopdf}
\usepackage{hyperref}
\usepackage{graphicx}
\usepackage{arydshln}

\newcolumntype{L}[1]{>{\raggedright\let\newline\\\arraybackslash\hspace{0pt}}m{#1}}
\newcolumntype{C}[1]{>{\centering\let\newline\\\arraybackslash\hspace{0pt}}m{#1}}
\newcolumntype{R}[1]{>{\raggedleft\let\newline\\\arraybackslash\hspace{0pt}}m{#1}}

\begin{document}

\newcommand\tal[1]{}
\newcommand\jon[1]{}
\newcommand\mbnote[1]{}

\newcommand\bb[1]{\textbf{#1}}

\newcommand{\denselist}{
\itemsep -2pt\topsep-8pt\partopsep-8pt
}

\pagestyle{headings}
\mainmatter

\title{Learning to Segment via Cut-and-Paste}

\author{Tal Remez, Jonathan Huang, Matthew Brown}
\institute{Google}

\maketitle

\begin{abstract}
This paper presents a weakly-supervised approach to object instance segmentation. Starting with known or predicted object bounding boxes, we learn object masks by playing a game of cut-and-paste in an adversarial learning setup. A mask generator takes a detection box and Faster R-CNN features, and constructs a segmentation mask that is used to cut-and-paste the object into a new image location. The discriminator tries to distinguish between real objects, and those cut and pasted via the generator, giving a learning signal that leads to improved object masks. We verify our method experimentally using Cityscapes, COCO, and aerial image datasets, learning to segment objects without ever having seen a mask in training. Our method exceeds the performance of existing weakly supervised methods \cite{rother2004grabcut,khoreva2016simple}, without requiring hand-tuned segment proposals, and reaches $90\%$ of supervised performance.
\vspace{-3mm}
\keywords{Instance segmentation, weakly-supervised, deep-learning.}
\end{abstract}

\begin{figure}[tb]
\centering
\begin{tabular}{c c c c c}
  \includegraphics[width=0.18\textwidth]{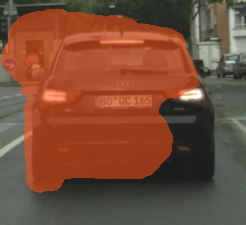} &
  \includegraphics[width=0.18\textwidth]{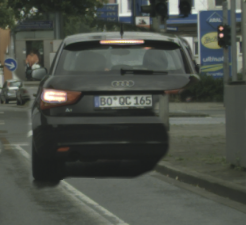} &
  \includegraphics[width=0.18\textwidth]{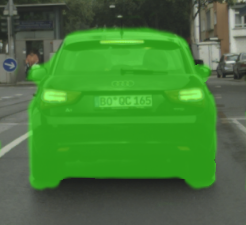} &
  \includegraphics[width=0.18\textwidth]{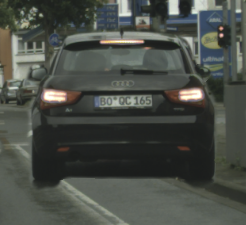} &
  \includegraphics[width=0.18\textwidth]{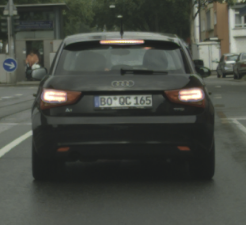}\\	
  (a) & (b) & (c) & (d) & (e)\\
  \end{tabular}
\caption{\footnotesize \textbf{Learning to Segment by Cut and Paste.} We iterate to learn accurate segmentation masks by trying to generate realistic images. A poor mask estimate (a) generates an unconvincing paste (b), while a good mask (c) results in a convincing one (d). Training a discriminator network to distinguish pasted from real images (e) creates a learning signal that encourages the generator to create better segmentations.}
\label{fig:teaser2}
\end{figure}

\begin{figure}[t]
\centering
    \begin{tabular}{ c }
    \includegraphics[width=0.99\textwidth]{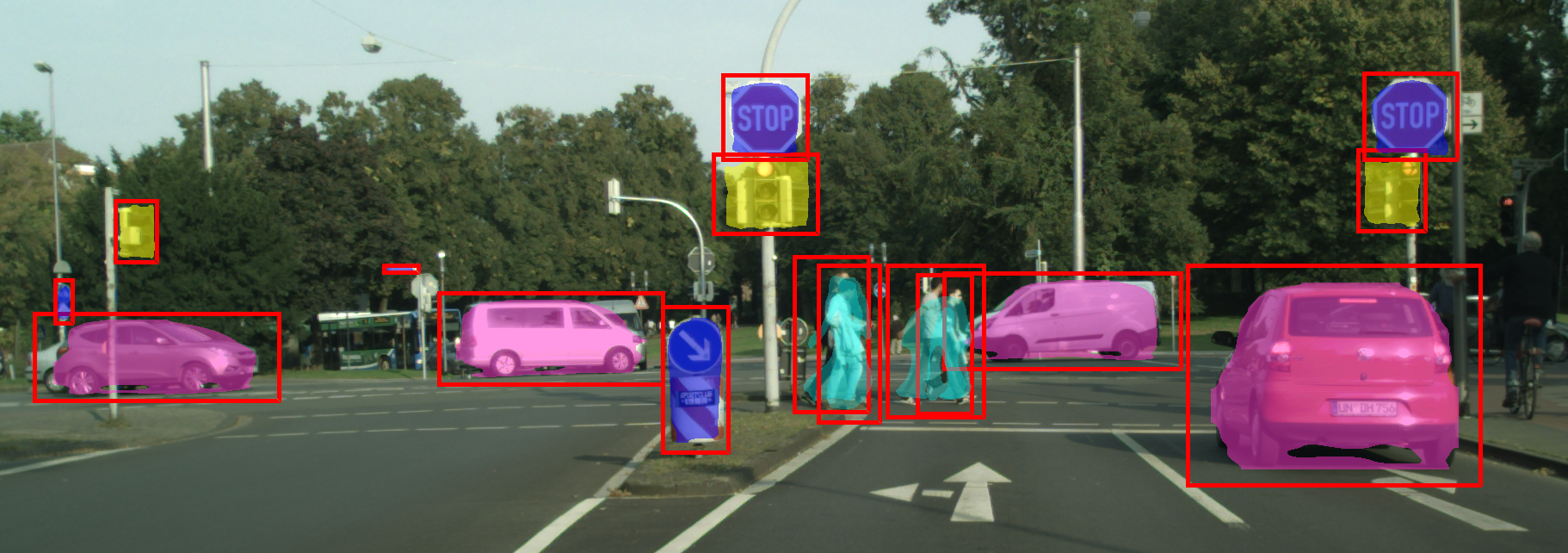}
    \end{tabular}
    \caption{\footnotesize Our method learns to segment objects without ever seeing ground truth masks and uses only bounding boxes as input.
    }
    \label{figure:cityscapes_example}
\end{figure}

\vspace{-2mm}
\section{Introduction}

Instance segmentation has seen much progress in recent years, with methods such as Mask R-CNN~\cite{he2017mask} now able to generate realistic masks, by building on the success of convolutional object detectors~\cite{ren2015faster,huang2016speed}.
Success has come at the cost of a significant labelling effort; the COCO segmentation dataset~\cite{lin2014microsoft} required around 40 person-years of labelling time for its 80 object categories.  

Modern object detection datasets have bounding boxes for up to 30k categories~\cite{krishna2017visual}. While still a considerable labelling effort, these bounding boxes can be generated roughly 10 times faster than the per-pixel segmentation masks required for fully supervised instance segmentation training. 
Moreover, labelling boxes has fixed complexity, whereas pixel-level labelling takes longer for objects with complex boundaries. In the COCO dataset,
for example, some complex object classes, such as `bicycle', are at best approximately labelled (see Figure 15 in~\cite{lin2014microsoft}). This motivates the question we address in this paper (Figure~\ref{figure:cityscapes_example}): can we learn instance
segmentation directly from bounding box data, and without ground truth masks?  We will propose a simple idea, which we call the ``cut-and-paste prior", to solve this problem (see Figure~\ref{fig:teaser2}). 

%
%
%
%
%


We are not the first to address the problem of box-supervised instance or semantic segmentation. Dai et al. propose a box-supervised method that uses an unsupervised candidate mask generator to create regression targets for semantic segmentation~\cite{dai2015boxsup}. The process is then iterated with the learned mask generator. Papandreou et al~\cite{papandreou2015weakly} propose a similar alternation, but using EM to calculate the pixel labels (E-step), and optimise the segmentation network parameters (M-step). ``Simple Does It'' by Khoreva et al.~\cite{khoreva2016simple} also follow Dai et al. in proposing synthetic regression targets. They experiment with using detection boxes alone, as well as Grabcut~\cite{rother2004grabcut} and MCG~\cite{pont2017multiscale} proposal generators, and also extend their approach to instance segmentation (using just 1 iteration in this case). Deepcut~\cite{rajchl2017deepcut} propose a modification to Grabcut using a CNN+CRF model. All of these approaches involve initialization and iteration between label estimation and generator training stages. Pathak et al.~\cite{pathak2015constrained} take a different approach, specifying hand-tuned constraints on the output label space (e.g., 75\% of pixels taking on the true label). Also related is ~\cite{wei2017object}, who grow object segmentation regions by sequentially discovering and erasing discriminative image regions, using classification rather than box supervision.

Our approach is qualitatively different from all these prior approaches. We require no segment proposals, pre-trained boundary detectors, or other hand-tuned initialization/constraints. Neither do we require iteration towards prediction and label consistency. Instead, our priors will be encapsulated in the structure of our generator/discriminator networks, and in our ``cut-and-paste" prior for object segmentation. 
The cut-and-paste prior encapsulates the basic idea that objects can move independently of their background. More precisely, objects may be cut out from one portion of an image, and pasted into another, and still appear realistic (see Figure~\ref{fig:teaser2}). With the help of a discriminator network to judge realism, we can use this process to provide a training signal for an instance segmentation network. 

We build on the successful approach of Generative Adversarial Networks (GANs)~\cite{goodfellow2014generative},
which have proved to be effective in modelling realistic images, e.g., hallucinating faces~\cite{karras2017progressive} and translating between image modalities~\cite{isola2017image}. 
However, rather than trying to generate images, we aim to generate segmentation masks. This allows us to use objective measures of performance (e.g., IoU against ground truth) for evaluation.
Related to our approach is the work of Luc et al.~\cite{luc2016semantic}, who also use an adversarial network to train a (semantic) segmentation algorithm. However, different to our approach, they use ground truth label maps as input to the discriminator. In our work we assume no such ground truth is available at training time.
Also related is the work of Hu et al.~\cite{hu2017learning} who use a partially supervised approach to generate object masks for a very large set of categories. 
They achieve this by joint learning using a set of fully supervised object classes and a larger set of box-only supervised classes, with a transfer function to map between box estimation and mask segmentation parameters. 
This seems to be a promising approach. 
However, in this work we focus on the unsupervised case, with only bounding boxes available for training.

Note that our approach of using cut-and-paste to form a loss function is \emph{not} the same as training data augmentation via cut-and-paste, e.g.,~\cite{dwibedi2017cut}, which takes existing masks and creates more training data out of it. This and related methods~\cite{georgakis2017synthesizing,alhaija2017augmented} do however exploit the same idea that image compositing can be used to create realistic imagery for training. They also note that object placement, lighting etc. are important; we revisit this topic in Sections~\ref{sec:wheretopaste1} and~\ref{sec:wheretopaste2}.


\subsection{Contributions}

The main contributions of this paper can be summarized as follows:
\vspace{-1mm}
\begin{itemize}
\item We propose and formalize 
a new \emph{cut-and-paste} adversarial
training scheme for box-supervised instance segmentation, which captures an intuitive prior, that objects can move independently of their background.  
\vspace{1mm}
\item We discuss the problem of identifying \emph{where} to paste
new objects in an image.
Even though objects are rarely truly independent of their background (e.g., cars do not typically appear in the middle of blue skies or on top of trees), we show that simple randomized
heuristics for selecting pasting locations
are surprisingly effective on real data.
\vspace{1mm}
\item 
Finally we showcase the success and generality of our approach by 
demonstrating that our method effectively learns to segment objects 
on a variety of datasets (street scenes, everyday objects, aerial imagery), without ever having access
to masks as supervision.  We also show
that our training method is stable and
yields models that outperform existing weakly supervised methods, reaching 90\% of supervised model performance.
\end{itemize}



\section{An Adversarial Formulation of the Cut and Paste Loss}
\label{section:adversarial_cut_and_paste}


\begin{figure}[t]
\centering
    \includegraphics[width=1.0\textwidth]{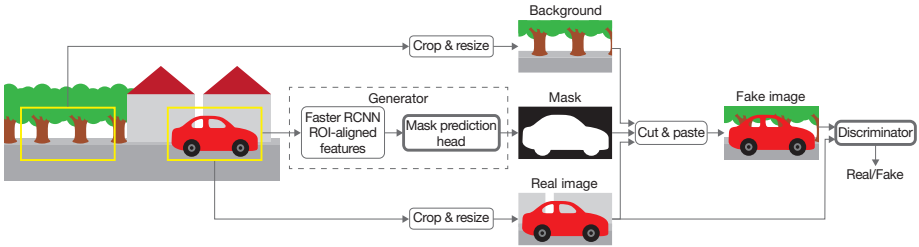}
    \caption{\textbf{Learning by cut-and-paste.}  
    A generator network receives a bounding box containing a car and predicts its mask. The discriminator alternately sees a cut+pasted car with a new background, or a real car image. Simultaneous training of generator and discriminator leads to improved object masks. Trainable blocks are outlined in {\bf bold}.}
    \label{figure:arch_overview}
\end{figure}



 An overview of our learning approach is shown in Figure~\ref{figure:arch_overview}.
We wish to train a model taking the form: $\mathcal{M} = G(X, \mathcal{B})$  that predicts an instance mask 
$\mathcal{M}$  
given an image $X$ 
and a bounding box $\mathcal{B}$ surrounding the instance of
interest.  
For simplicity we will ignore classes and typically assume that instances are of the same
class (e.g., `person' or `car'), training an independent model per
class.  
Intuitively, we would
like to assign a low loss to a predicted mask if copying the
pixels from the mask $\mathcal{M}$ and pasting into a new part of the image $X$
yields a \emph{plausible} image patch and high loss otherwise (see Figure~\ref{fig:teaser2}).




In order to measure the notion of ``plausibility'',
we use a GAN, viewing the function $G$ as a \emph{generator}.
Given a generated mask $\mathcal{M}$, we synthesize a new image patch $F$ by compositing image $X_{\mathcal{B}}$ from bounding box $\mathcal{B}$ with a new background image $X_{\mathcal{B}'}$ from location $\mathcal{B}'$ (typically in the same image):

\vspace{-4mm}
\begin{align}
    F = \mathcal{M}X_{\mathcal{B}}+(1-\mathcal{M})X_{\mathcal{B}'}.
\label{eqn:compositing}
\end{align}


%


\noindent The fake image $F$ is fed to a second model, the \emph{discriminator}, whose job is to distinguish whether $F$ is real or synthesized.  We now simultaneously train the discriminator to distinguish reals from fakes and the generator to make the discriminator's error rate as high as possible.
More formally, we maximize with respect to parameters of the discriminator
$D$ and minimize with respect to parameters of the generator $G$ in the following loss function:



\vspace{-4mm}
  \begin{align}
    \mathcal{L}_{CPGAN} & =  \mathbb{E} \; \log D(X_{\mathcal{B}})\, + \,\log(1 - D(F)).
  \end{align}

We refer to this as the \emph{cut-and-paste} loss, since it aims to align real images and their cut-and-pasted counterparts. 
Note that the fake image $F$ is a function of the generator $G$ via the mask $\mathcal{M} = G(X, \mathcal{B})$, as specified in Equation~\ref{eqn:compositing}. 
The expectations are over $(X,\mathcal{B})\sim p_{data}$ being the input set of images and bounding boxes, with $\mathcal{B}'$ drawn randomly as described in the Section \ref{sec:wheretopaste1} below. 

Over training iterations, the hope is that the only way that the generator can successfully ``fool'' the discriminator is by generating correct masks.  
We now discuss several critical stepping stones to get such a model to train effectively.


\subsection{Where to Paste}
\label{sec:wheretopaste1}

\begin{figure}[tb]
\centering
\begin{tabular}{ c c }
       
        \includegraphics[width=0.49\textwidth]{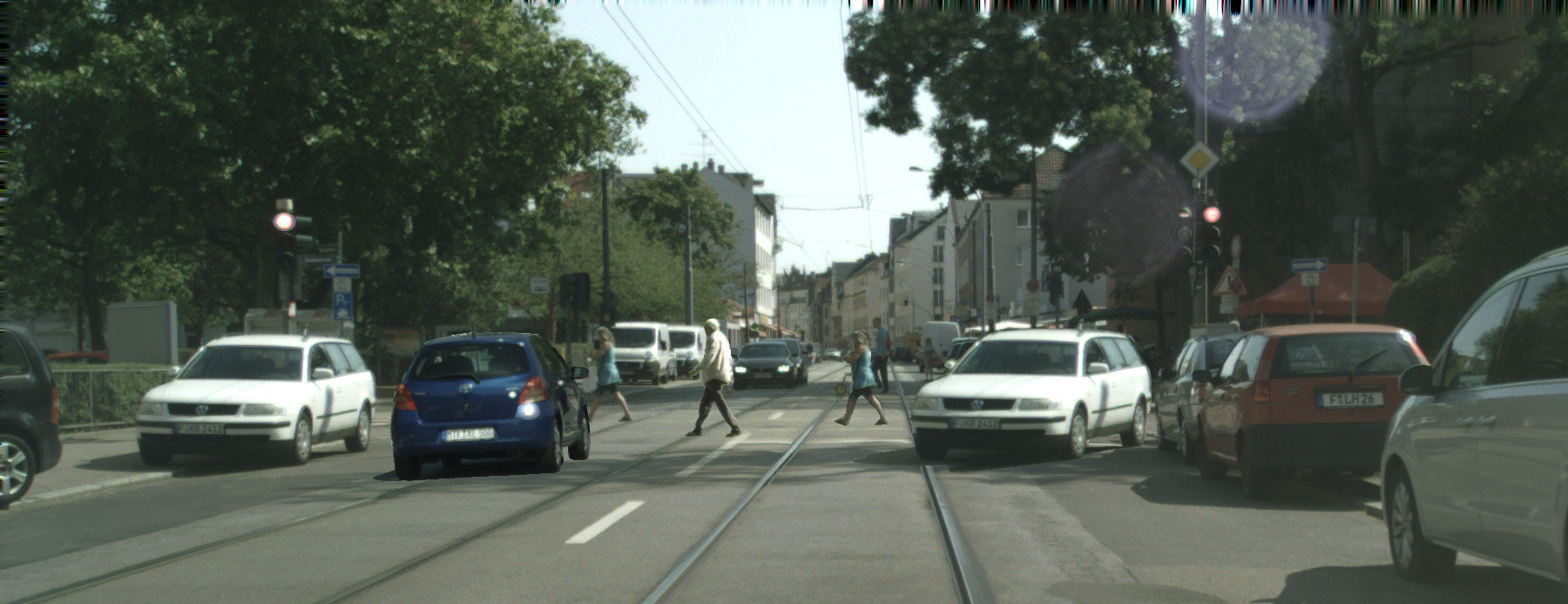} &
        \includegraphics[width=0.49\textwidth]{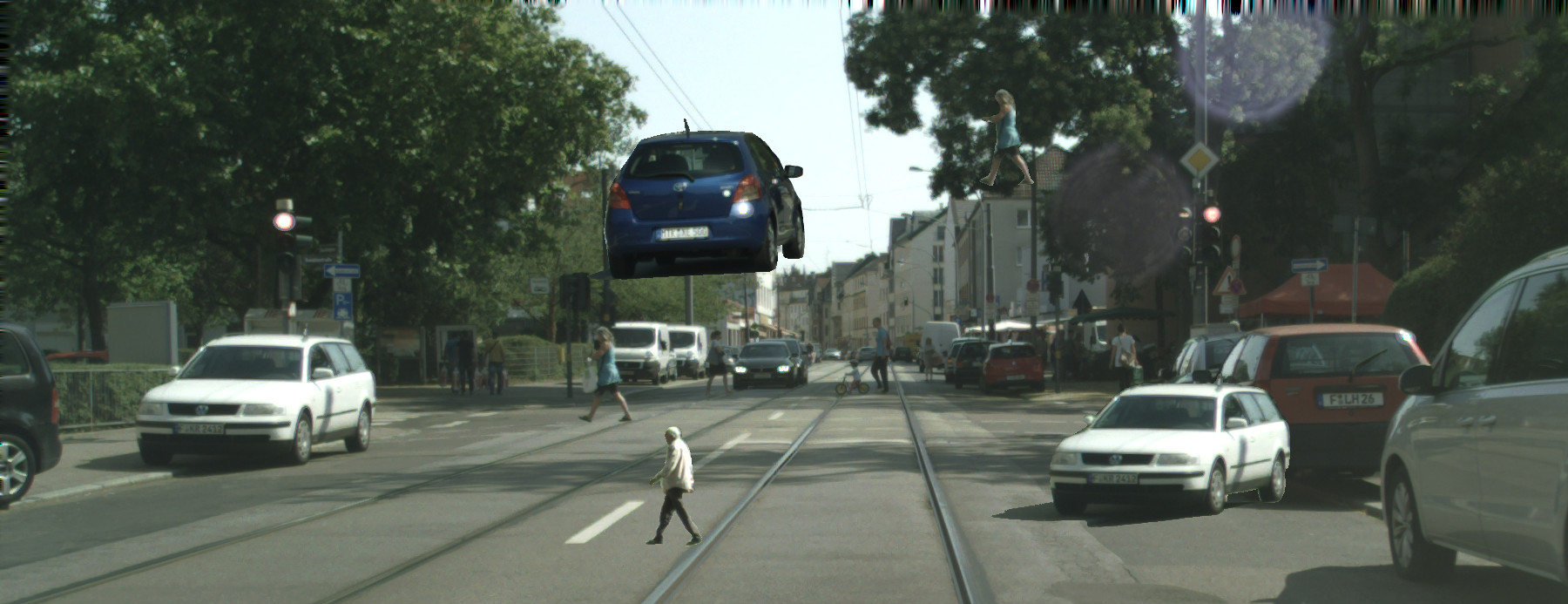} \\
        (a) & (b) \\
        
\end{tabular}
\caption{\textbf{Cut-and-paste locations.} A few objects have been cut-and-pasted to new locations in the original image. In (a) they were pasted along the same scanline as their original position, making it harder to tell them apart (can you spot them?); In (b) they were pasted at random positions making it much easier.}
\label{figure:cut_paste_location}
\end{figure}

The choice of where to paste an object to generate a realistic looking result is clearly important for human observers, e.g., see Figure~\ref{figure:cut_paste_location}.
It is also data dependent. For example, buildings may appear at any $(x,y)$ location in our aerial imagery with equal probability (Figure \ref{figure:aerial_overlays}), but realistic pedestrian placement and scale is highly constrained in street scenes. Whilst sophisticated pasting strategies might be devised, we find that good results can still be obtained using simple ones. In this work we experiment with two main pasting strategies: 
1) Uniform pasting: paste anywhere into the same image, taking care not to overlap the same object class, 2) Depth sensitive pasting: we take care to preserve the correct scale when pasting using knowledge of the scene geometry.
We discuss this further in the experiments reported in Section~\ref{sec:wheretopaste2} on the Cityscapes, COCO, and aerial imagery datasets.

\subsection{Avoiding Degenerate Solutions}
\label{sec:avoiding_degenerate_solutions}

Our learning objective is based on realism in the pasted result; 
this strategy usually leads to good solutions, but there are a few degenerate cases. For example, realistic images can be generated by choosing all of the pixels or none of the pixels in the bounding box (though in the latter case this doesn't contain the object).
Also, some objects are modular and part of the object can be pasted and still give a realistic image. We examine each of these cases in turn. 

The first case (generator marks all pixels as foreground) can be mitigated by giving the discriminator a larger viewport than the region into which the generator pastes. Giving the discriminator a small band of context around the pasted object (typically 10\% of box width) allows for easy identification of this failure mode, as the background will change abruptly at the bounding box borders.\footnote{Note that this strategy will fail in cases where source and destination backgrounds are identical, e.g., pasting an airplane from one blue sky to another, but these cases are rare for most classes.}

If the generator decides to label none of the pixels as belonging to the object,
%
%
the resulting fake image will look realistic, but will not contain the object of interest. This case should be automatically solved in our framework, since the discriminator will expect to see the object present. 
However, we found that adding an explicit classification loss significantly aids stability and improves performance in some cases.
To this end we add an additional classification network $\textit{D}_{CLS}$ which explicitly encourages the model to ensure that the object of interest is really present (see Figure \ref{figure:aerial_plots}(b)).
One way to think about this new loss is that our generator is now trying to fool two discriminators: one that had been trained on a previous classification task (and is frozen), and another that is training and evolving with the generator. This gives an additional classification loss term for the generator:


\vspace{-4mm}
  \begin{align}
  \mathcal{L}_{CLS} & = \mathbb{E} \; \log(1-D_{CLS}(F)).
  \end{align}


A final failure mode can occur if the generator chooses to paste some sub-part of an object that may still be realistic in isolation, e.g., part of a building or other modular structure. This is to some extent addressed by the classification loss $\mathcal{L}_{CLS}$, which favours complete objects being pasted. However, we also explore a complementary 
\emph{cut loss} to address this, as described below.

\subsection{Adversarial Cut Loss}

\begin{figure}[tb]
\centering
\small
\begin{tabular}{c c c c c c}
  \includegraphics[width=0.14\textwidth]{Figures/teaser2/badmask.png} &
  \includegraphics[width=0.14\textwidth]{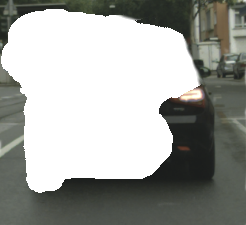} &
  \includegraphics[width=0.14\textwidth]{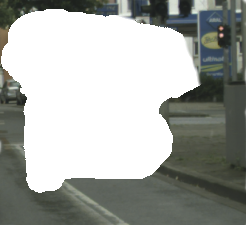} &
  \includegraphics[width=0.14\textwidth]{Figures/teaser2/goodmask.png} &
  \includegraphics[width=0.14\textwidth]{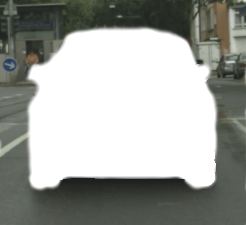} &
  \includegraphics[width=0.14\textwidth]{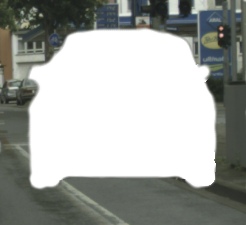}\\	
  (a) & (b) & (c) & (d) & (e) & (f)\\
  \end{tabular}
\caption{\textbf{Cut Loss Illustration.} A poor mask (a) leaves object parts behind, hence there is a difference when applied to the input (b) or random image (c). A good mask (d) gives indistinguishable results when applied to the input (e) or random image (f).}
\label{fig:cut_loss_demo}
\end{figure}

In the basic version of our model, a generator learns to identify the correct subset of image pixels such that pasting those pixels into a new image results in a believable result. However, for certain objects, such as modular or repeated structures, it is possible that copying a sub-part of the object results in a plausible image. One example that we have encountered is buildings, where sub-parts often resemble realistic buildings (Figure~\ref{figure:aerial_overlays}). In these cases, the part of the scene left behind after the cut operation will often contain parts of the object.

To mitigate this effect, we set up a secondary adversarial game, that takes as input the background we get after cutting out our object $\tilde{X}_{\mathcal{B}}$, and compares this to the same cut applied elsewhere in the scene, $\tilde{F}$.  Specifically, we have that
$\tilde{X}_{\mathcal{B}} = (1-\mathcal{M})X_\mathcal{B}$ and $\tilde{F} = (1-\mathcal{M})X_{\mathcal{B}'}$.
In this new adversarial game, our mask generator must now also fool a new  discriminator, $D_{CUT}$, leading to an additional loss term:

 
 
 
    
\vspace{-4mm}    
\begin{align}
    \mathcal{L}_{CGAN} = \mathbb{E} \; \log D((1-\mathcal{M})X_{\mathcal{B}})\, + \,\log(1 - D((1-\mathcal{M})X_{\mathcal{B}'})),
  \end{align}

\noindent which we refer to as our ``cut" loss, as it penalizes differences between the cut operation on real and background imagery (see Figure~\ref{fig:cut_loss_demo}).

\subsection{Overall Loss Function} 
 
Our overall loss function is the sum of cut+paste, classification and cut losses:

\vspace{-4mm}
\begin{align}
  \mathcal{L} = \mathcal{L}_{CPGAN} + w_{cls}\mathcal{L}_{CLS} + w_{cut}\mathcal{L}_{CGAN}.
\end{align}

\noindent In practice, we use a LSGAN~\cite{mao2017least} formulation, which converts min/max optimization of GAN loss terms of the form $\mathcal{L} = \mathbb{E} \; \log (1-D(X)) \, + \, \log (1 - D(G(X)))$ into separate optimizations for the discriminator and generator:

\vspace{-4mm}
\begin{align}
\min_{D} \; \mathbb{E} \; (D(G(X))^2 + (D(X)-1)^2, \;\; \min_{G} \; \mathbb{E} \; (D(G(X)-1)^2.
\end{align}

\section{Architecture}

There are three modules in our model: (1) the generator, which predicts a mask, (2) the cut-and-paste module, which produces a ``fake patch'' given the predicted mask, and (3), the discriminator,
which distinguishes between real and fake patches, see Figure \ref{figure:arch_overview}.
In the following, we describe the architecture for each of these modules
that we have used in our experiments.

\begin{table}[tb]
\centering
\scriptsize
\begin{tabular}{ c c }
\textbf{Generator} & \textbf{Discriminator} \\
\begin{tabular}{ | l | l | }
  \hline
  Output size               & Layer \\ 
  \hline
  $7\!\!\times\!\!7\!\!\times\!\!2048$      & Input, ROI-aligned features \\
  $7\!\!\times\!\!7\!\!\times\!\!256$       & Conv, $1\!\!\times\!\!1\times256$, stride 1 \\
  $7\!\!\times\!\!7\!\!\times\!\!256$       & Conv, $3\!\!\times\!\!3\!\!\times\!\!256$, stride 1 \\
  $14\!\!\times\!\!14\!\!\times\!\!256$     & Bilinear upsampling \\  
  $14\!\!\times\!\!14\!\!\times\!\!256$     & Conv, $3\!\!\times\!\!3\!\!\times\!\!256$, stride 1 \\
  $28\!\!\times\!\!28\!\!\times\!\!256$     & Bilinear upsampling \\
  $28\!\!\times\!\!28\!\!\times\!\!256$     & Conv, $3\!\!\times\!\!3\!\!\times\!\!256$, stride 1 \\
  $28\!\!\times\!\!28\!\!\times\!\!1$       & Conv, $3\!\!\times\!\!3\!\!\times\!\!1$, stride 1 \\
  $28\!\!\times\!\!28\!\!\times\!\!1$       & Sigmoid \\
  \hline
\end{tabular} &
\qquad
\begin{tabular}{c}
\begin{tabular}{ | l | l | }
  \hline
  Output size               & Layer \\ 
  \hline
  $34\!\!\times\!\!34\!\!\times\!\!3$       & Input image patch \\
  $32\!\!\times\!\!32\!\!\times\!\!64$      & Conv, $3\!\!\times\!\!3\!\!\times\!\!64$, stride 1, valid \\
  $15\!\!\times\!\!15\!\!\times\!\!128$     & Conv, $3\!\!\times\!\!3\!\!\times\!\!128$, stride 2, valid \\
  $7\!\!\times\!\!7\!\!\times\!\!256$       & Conv, $3\!\!\times\!\!3\!\!\times\!\!256$, stride 2, valid \\
  $3\!\!\times\!\!3\!\!\times\!\!512$       & Conv, $3\!\!\times\!\!3\!\!\times\!\!512$, stride 2, valid \\  
  $4608$                    & Flatten \\
  $2$                       & Fully connected \\
  $2$                       & Softmax \\
  \hline
\end{tabular} \\
\begin{tabular}{c}
\\
\end{tabular}
\end{tabular} \\
\end{tabular}
\caption{\footnotesize \textbf{Generator and discriminator architectures.}  Our generator takes ROI-aligned features from a Faster R-CNN  detector and applies a mask prediction head similar to that used in Mask R-CNN~\cite{he2017mask}. Our discriminator is applied directly on $34\!\times\!34$ image patches.  After each convolution we use 
ReLU nonlinearities for the generator and Leaky ReLUs (with $\alpha=0.2$) for the discriminator.}
\label{table:generator_and_discriminator_arch}
\vspace{-3mm}
\end{table}

\noindent\textbf{Generator:} 
\label{section:architecture}
Our generator is similar to that of Mask R-CNN~\cite{he2017mask}. A ResNet-50 backbone is used to extract ROI-aligned features and a mask prediction head is applied to these features.
%
Our mask prediction head is described in Table \ref{table:generator_and_discriminator_arch}, and is comprised of a series of convolutions, bilinear upsampling operations, and a Sigmoid nonlinearity resulting in a $28\times 28$ mask output. We find that using corner-aligned bilinear upsampling generally provides better results than transposed convolutions and nearest neighbour upsampling layers.

\noindent\textbf{Cut-and-Paste Module:}
%
%
We implement the cut-and-paste operation using standard alpha compositing (Equation~\ref{eqn:compositing}). The inferred mask is typically at a lower resolution than the foreground and background images, so we downsample to the mask resolution before compositing. Note that careful sampling in this step is critical, as convolutional networks can easily detect any aliasing or blurring artifacts, which are easy indicators that an image is fake.
As explained in Section \ref{sec:avoiding_degenerate_solutions}, we allow the discriminator a larger viewport than the original mask size, therefore our $28\times28$ masks are padded with 3 pixels of zeros on each side.


%
%


\noindent\textbf{Discriminator:}
Our discriminator receives an $N\!\times\!N$ image patch as input, and predicts whether the
given patch is real or fake. Our discriminator architecture is presented in Table~\ref{table:generator_and_discriminator_arch}, and is comprised of a series of valid convolutions (convolutions without padding) followed by a fully connected layer and a Softmax. 

\noindent\textbf{Training Procedure:}
Our models are implemented in TensorFlow \cite{abadi2016tensorflow} and are trained using a batch size of 4 instances for the generator and 8 instances for the discriminator (4 real and 4 fake). We use the Adam optimizer~\cite{kingma2014adam} 
with learning rate of $5\cdot10^{-5}$, $\beta_1=0.9$, $\beta_2=0.999$, and $\epsilon=10^{-8}$. We train for $1$ million iterations, alternating optimization equally between generator and discriminator.
Our supervised model is trained similarly but with a cross-entropy loss on the ground truth masks. 
%
The backbone generating the features for our generator was pretrained on the COCO detection challenge data 
and is held frozen through training. The rest of the generator and discriminator layers are initialized using random Xavier initialization \cite{glorot2010understanding}. CityScapes and COCO training data are augmented by adding random horizontal flips.

\section{Experiments}
In this section we present the results of experiments using street scenes (Cityscapes), common objects (COCO) and aerial image datasets. Overall results (Tables \ref{table:cityscapes_results} and \ref{table:coco_results}) indicate that 
our models are competitive
or better than other weakly supervised baselines.
We also investigate some of the strengths and failure modes of our approach, including analysing dataset specific performance, effect of pasting strategies, settings for loss hyperparameters, and the effect of data scaling.



%


\subsection{Evaluation Methodology and Baselines.}
\label{section:eval_methodology}

%
We compare our proposed approach (which we will refer to in below tables as
{\bf Cut\&Paste}) to a few baseline methods, all of which take as input (1) an image and (2) a bounding box surrounding the
instance to be segmented, and output a segmentation mask.
The simplest baseline strategy (which we call {\bf Box}) is to simply declare 
all pixels within the given ground truth bounding box to be the foreground/object. Since bounding boxes are tight around the objects in the datasets that we use, this is often a reasonable guess, assuming that no additional information is available.
Another classic baseline is the {\bf GrabCut}~\cite{rother2004grabcut} algorithm. We use 5 iterations of the OpenCV implementation, guiding with a central foreground rectangle 40\% of the box size if the initial iterations return a zero-mask.
%

We also evaluate the performance of the recent {\bf Simple Does It} approach
by Khoreva et al.,~\cite{khoreva2016simple} by running their publicly available pretrained instance segmentation model $DeepLab_{BOX}$, which was trained on PASCAL VOC \cite{pascal-voc-2012} and COCO.

%
In addition to these baselines, we also train a fully supervised version of our model (called {\bf FullySupervised}), which uses the same architecture as our generator, but is trained using cross entropy loss against ground truth masks. This gives us an idea of the best performance we should expect from our weakly supervised methods.


For methods outputting low-resolution masks (this includes {\bf Cut\&Paste}, {\bf FullySupervised}, and {\bf Simple Does It}), we resize their masks using bicubic interpolation back to the original image resolution prior to evaluation.


In contrast to typical generative models of images based on GANs, we can evaluate our method based on objective measures. We present results in this section in terms of the mean intersection-over-union (\emph{mIoU}) measure,
a commonly used metric for segmentation. Since our bounding boxes are assumed to be given, 
we refrain from presenting average precision/recall based measures such as those used by the COCO dataset since they depend on the detected boxes. 


\subsection{CityScapes}


\begin{table}[tb]
\begin{center}
\footnotesize
\begin{tabular}{ | l | C{1.5cm} | C{1.5cm} | C{1.5cm} | C{1.5cm} | C{1.5cm} |}
  \hline
        Method                                                  & Car       & Person    & Traffic-light     & Traffic-sign    \\ \hline
  \multirow{1}{*}{\bf Box}                                          & 0.62      & 0.49      & 0.76              & 0.79           \\ \hline        
  
  \multirow{1}{*}{\bf GrabCut \cite{rother2004grabcut}}            & 0.62      & 0.50      & 0.64              & 0.65          \\ \hline
  
  
  \multirow{1}{*}{\bf Simple Does It \cite{khoreva2016simple}}      & \bb{0.68} & 0.53      & 0.60              & 0.51          \\ \hline
  \multirow{1}{*}{\bf Cut\&Paste (Ours)}                            & 0.67      & \bb{0.54} & \bb{0.77}         & \bb{0.79}         \\ \hline \hline
  \multirow{1}{*}{\bf FullySupervised}                              & 0.80      & 0.61      & 0.79              & 0.81           \\ \hline 

\end{tabular}
\end{center}
\vspace{-3mm}
\caption{
\textbf{mIOU performance on Cityscapes}}
\label{table:cityscapes_results}
\end{table}

\begin{figure}[tb]
\scriptsize
\centering
\begin{tabular}{ c@{\hspace{0.005\textwidth}}c@{\hspace{0.005\textwidth}}c}

\begin{tabular}{ c@{\hspace{0.005\textwidth}}c@{\hspace{0.005\textwidth}}c@{\hspace{0.005\textwidth}}c}
    Image & GT & \cite{khoreva2016simple} & Ours \\
    \includegraphics[width=0.93cm,height=1.9cm]{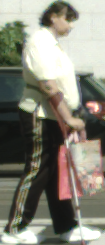} &
    \includegraphics[width=0.93cm,height=1.9cm]{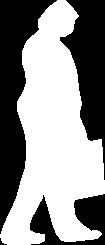} &
    \includegraphics[width=0.93cm,height=1.9cm]{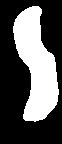} &
    \includegraphics[width=0.93cm,height=1.9cm]{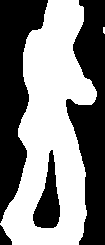} \\ [-0.25ex]
    
    \includegraphics[width=0.93cm,height=0.93cm]{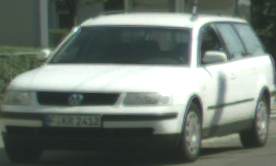} &
    \includegraphics[width=0.93cm,height=0.93cm]{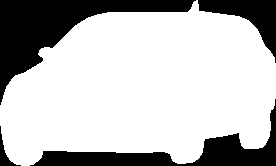} &
    \includegraphics[width=0.93cm,height=0.93cm]{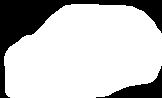} &
    \includegraphics[width=0.93cm,height=0.93cm]{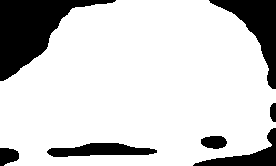} \\[-0.25ex]
\end{tabular} & 
\begin{tabular}{ c@{\hspace{0.005\textwidth}}c@{\hspace{0.005\textwidth}}c@{\hspace{0.005\textwidth}}c}         Image & GT & \cite{khoreva2016simple} & Ours \\
    \includegraphics[width=0.93cm,height=1.9cm]{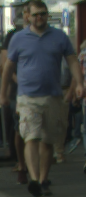} &
    \includegraphics[width=0.93cm,height=1.9cm]{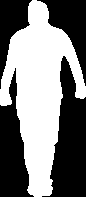} &
    \includegraphics[width=0.93cm,height=1.9cm]{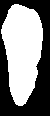} &
    \includegraphics[width=0.93cm,height=1.9cm]{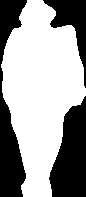}  \\ [-0.25ex]

    \includegraphics[width=0.93cm,height=0.93cm]{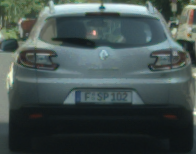} &
    \includegraphics[width=0.93cm,height=0.93cm]{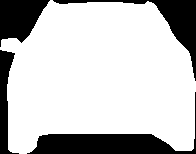} &
    \includegraphics[width=0.93cm,height=0.93cm]{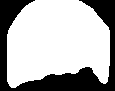} &
    \includegraphics[width=0.93cm,height=0.93cm]{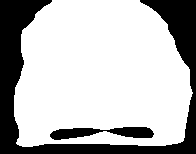} \\[-0.25ex]
\end{tabular} &
\begin{tabular}{ c@{\hspace{0.005\textwidth}}c@{\hspace{0.005\textwidth}}c@{\hspace{0.005\textwidth}}c} 
    Image & GT & \cite{khoreva2016simple} & Ours \\
    \includegraphics[width=0.93cm,height=0.93cm]{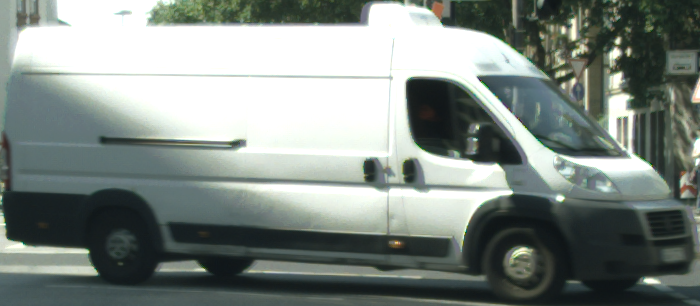} &
    \includegraphics[width=0.93cm,height=0.93cm]{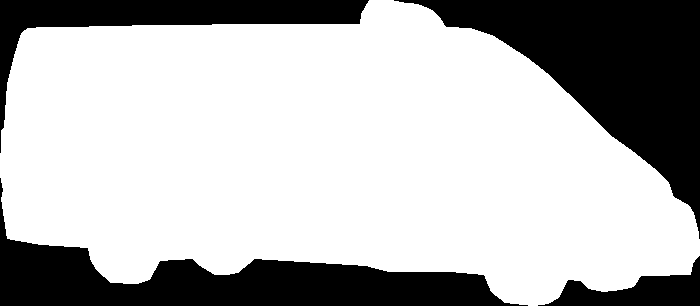} &
    \includegraphics[width=0.93cm,height=0.93cm]{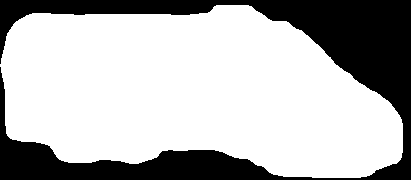} &
    \includegraphics[width=0.93cm,height=0.93cm]{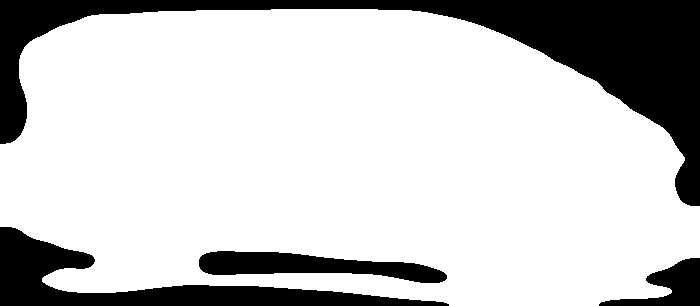} \\[-0.37ex]
    
    \includegraphics[width=0.93cm,height=0.93cm]{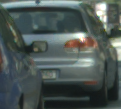} &
    \includegraphics[width=0.93cm,height=0.93cm]{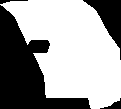} &
    \includegraphics[width=0.93cm,height=0.93cm]{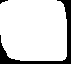} &
    \includegraphics[width=0.93cm,height=0.93cm]{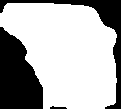} \\[-0.37ex]        
    
    \includegraphics[width=0.93cm,height=0.93cm]{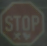} &
    \includegraphics[width=0.93cm,height=0.93cm]{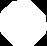} &
    \includegraphics[width=0.93cm,height=0.93cm]{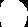} &
    \includegraphics[width=0.93cm,height=0.93cm]{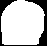} \\[-0.37ex]    
    
\end{tabular}
\end{tabular}
\caption{\textbf{Cityscapes mask comparison.} From left to right: the original image, the ground truth mask (GT), the mask predicted by Simple Does It \cite{khoreva2016simple}, and our mask.}
\label{figure:cityscapes_results}
\end{figure}

\begin{figure}[t]
\centering
\scriptsize
\centering
\begin{tabular}{ c@{\hspace{0.003\textwidth}}c@{\hspace{0.003\textwidth}}c@{\hspace{0.003\textwidth}}c@{\hspace{0.003\textwidth}}c@{\hspace{0.003\textwidth}}c@{\hspace{0.003\textwidth}}c@{\hspace{0.003\textwidth}}c@{\hspace{0.003\textwidth}}c@{\hspace{0.003\textwidth}}c@{\hspace{0.003\textwidth}}c@{\hspace{0.003\textwidth}}c@{\hspace{0.003\textwidth}}c@{\hspace{0.003\textwidth}}c@{\hspace{0.003\textwidth}}c@{\hspace{0.003\textwidth}}c@{\hspace{0.003\textwidth}}c}
        \includegraphics[height=1.3cm]{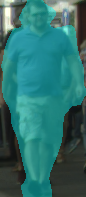} &   
        \includegraphics[height=1.3cm]{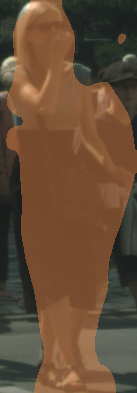} &
        \includegraphics[height=1.3cm]{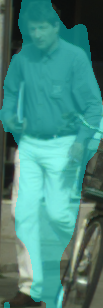} &
        \includegraphics[height=1.3cm]{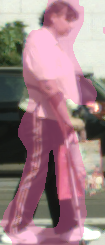} &
        \includegraphics[height=1.3cm]{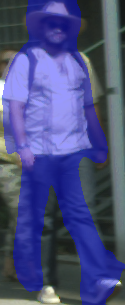} &
        \includegraphics[height=1.3cm]{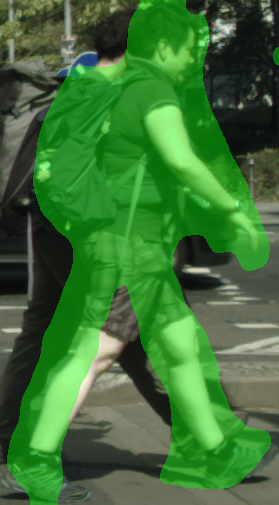} &
        \includegraphics[height=1.3cm]{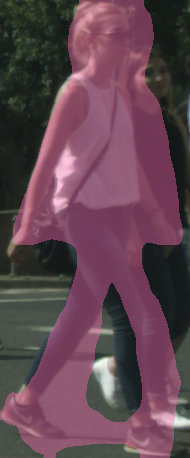} &
        \includegraphics[height=1.3cm]{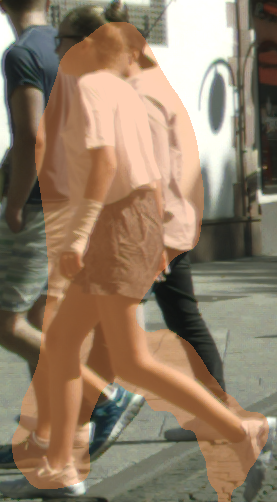} &
        \includegraphics[height=1.3cm]{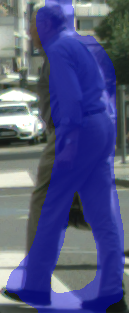} &
        \includegraphics[height=1.3cm]{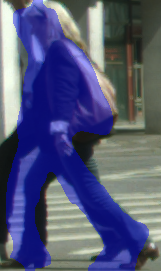} &
        \includegraphics[height=1.3cm]{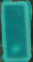} &
        \includegraphics[height=1.3cm]{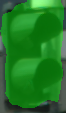} &
        \includegraphics[height=1.3cm]{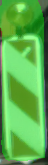} &
        \includegraphics[height=1.3cm]{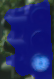} &
        \includegraphics[height=1.3cm]{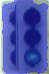} &
        \includegraphics[height=1.3cm]{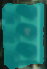} &        
        \includegraphics[height=1.3cm]{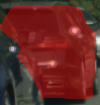} \\ [-0.5ex]
\end{tabular}        
\begin{tabular}{ c@{\hspace{0.003\textwidth}}c@{\hspace{0.003\textwidth}}c@{\hspace{0.003\textwidth}}c@{\hspace{0.003\textwidth}}c@{\hspace{0.003\textwidth}}c@{\hspace{0.003\textwidth}}c}        

        \includegraphics[height=1.2cm]{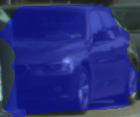} &   
        \includegraphics[height=1.2cm]{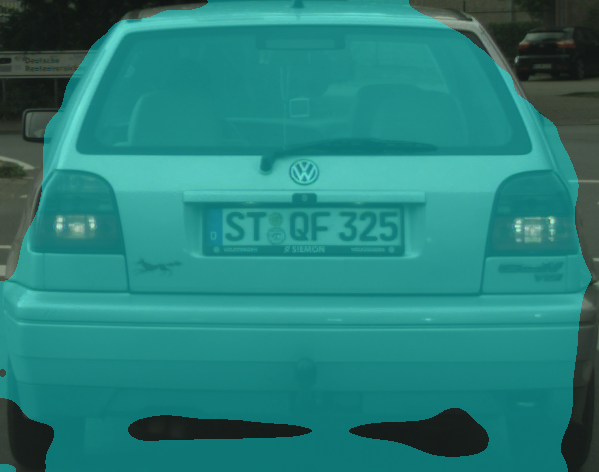} &
        \includegraphics[height=1.2cm]{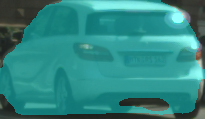} &
        \includegraphics[height=1.2cm]{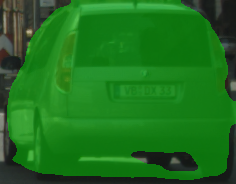} &
        \includegraphics[height=1.2cm]{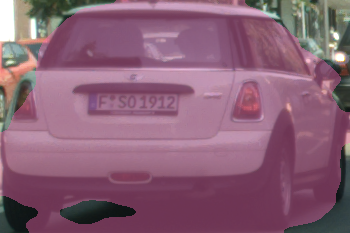} &
        \includegraphics[height=1.2cm]{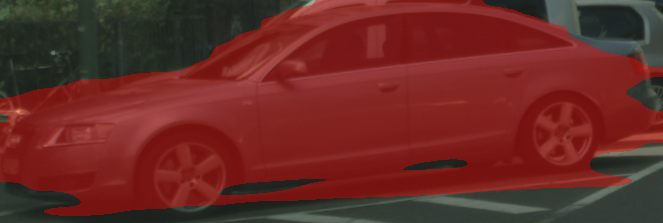} \\ [-0.5ex]
\end{tabular}
\begin{tabular}{ c@{\hspace{0.0035\textwidth}}c@{\hspace{0.0035\textwidth}}c@{\hspace{0.0035\textwidth}}c@{\hspace{0.0053\textwidth}}c@{\hspace{0.0035\textwidth}}c@{\hspace{0.0035\textwidth}}c}        
        \includegraphics[height=1.2cm]{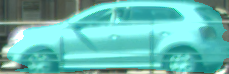} &
        \includegraphics[height=1.2cm]{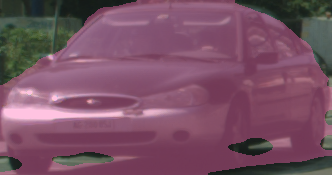} &
        \includegraphics[height=1.2cm]{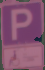} &   
        \includegraphics[height=1.2cm]{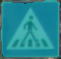} &
        \includegraphics[height=1.2cm]{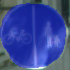} &
        \includegraphics[height=1.2cm]{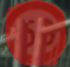} &
        \includegraphics[height=1.2cm]{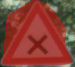} \\
\end{tabular}
\caption{\textbf{Cityscapes examples.} Masks produced by our method.}
\label{figure:cityscapes_results_overlay}
\end{figure}

The CityScapes dataset~\cite{Cordts2016Cityscapes} 
consists of densely annotated imagery of
street scenes from cameras mounted on a car. Images are usually wide enough that it is easy to find plausible pasting positions for fake objects, so one would expect our method to perform well on this kind of data.

To prepare the CityScapes data for training our models,
we separate the official training set into a training set and a development set (the latter containing the sequences of Aachen, Bremen and Bochum), using
the official validation set as a test set for all methods. 
We extract instance segmentation masks from the fine-grained annotated ``left-camera'' images for four classes: 
``car'', ``person'', ``traffic-light'' and ``traffic-sign''\footnote{``Traffic-light'' and ``Traffic-sign'' instance segmentation masks are not provided with the dataset, but semantic segmentation masks \emph{are} provided; thus to extract masks, we consider each connected component of these classes as a separate instance.},
removing any
cars or people smaller than 100 pixels along either axis, 
and any traffic lights or signs smaller than 25 pixels. 
Ground truth instance segmentations are used for evaluation, and for training the supervised version of our model. For the box-supervised version, we use a combination of ground truth bounding boxes from the $2,975$ annotated images, and additional bounding boxes generated by running a Faster R-CNN object detector\footnote{The detector is pretrained on the COCO dataset, the model can be found in TensorFlow Object Detection API model zoo:
\url{https://github.com/tensorflow/models/blob/master/research/object_detection/}}
on the $89,240$ unannotated images in the \texttt{leftImg8bit\_sequence} set.





Our results, shown in Table~\ref{table:cityscapes_results}, demonstrate that
across all four classes, we are consistently better than the {\bf Box}
and  {\bf GrabCut} baselines. Note that {\bf Box} performs suprisingly well on some of these classes, notably signs and traffic lights, for which the ground truth bounding box is typically already a good fit. We also outperform the {\bf Simple Does It} approach and are within 90\% of our fully supervised baseline on all but the ``Car'' class.
Figure \ref{figure:cityscapes_results} shows a qualitative comparison between
masks generated by our method and those by {\bf Simple Does It}. Typically the masks from both methods are comparable in quality, except in the case of people where our {\bf Cut\&Paste} method performs noticeably better, especially in fine details such as arms and legs. 
Figure \ref{figure:cityscapes_results_overlay} presents more examples of our masks.
These results used the $\mathcal{L}_{CPGAN}$ loss, with zero weight to cut and classification loss terms ($w_{cls}=w_{cut}=0$). See Section~\ref{sec:aerial_results} for discussion and results of loss term weightings. All methods were evaluated on images at $600\times 1200$ resolution.

\begin{figure}[tb]
\centering
\footnotesize
\begin{tabular}{ r@{\hspace{0.005\textwidth}}c@{\hspace{0.005\textwidth}}c@{\hspace{0.005\textwidth}}c@{\hspace{0.005\textwidth}}c@{\hspace{0.005\textwidth}}c@{\hspace{0.005\textwidth}}c@{\hspace{0.005\textwidth}}c@{\hspace{0.005\textwidth}}c@{\hspace{0.005\textwidth}}c@{\hspace{0.005\textwidth}}c }

    \vspace{0.5mm}Real &
    \raisebox{-.5\height}{\includegraphics[width=1.0cm]{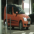}} &
    \raisebox{-.5\height}{\includegraphics[width=1.0cm]{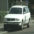}} &
    \raisebox{-.5\height}{\includegraphics[width=1.0cm]{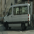}} &
    \raisebox{-.5\height}{\includegraphics[width=1.0cm]{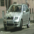}} &
    \raisebox{-.5\height}{\includegraphics[width=1.0cm]{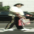}} &
    \raisebox{-.5\height}{\includegraphics[width=1.0cm]{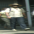}} &
    \raisebox{-.5\height}{\includegraphics[width=1.0cm]{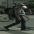}} &
    
    
    \raisebox{-.5\height}{\includegraphics[width=1.0cm]{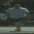}} &
    \raisebox{-.5\height}{\includegraphics[width=1.0cm]{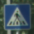}} &
    \raisebox{-.5\height}{\includegraphics[width=1.0cm]{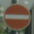}} \\

    \vspace{0.5mm}Fake &
    \raisebox{-.5\height}{\includegraphics[width=1.0cm]{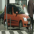}} &
    \raisebox{-.5\height}{\includegraphics[width=1.0cm]{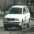}} &   
    \raisebox{-.5\height}{\includegraphics[width=1.0cm]{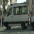}} &
    \raisebox{-.5\height}{\includegraphics[width=1.0cm]{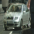}} &
    \raisebox{-.5\height}{\includegraphics[width=1.0cm]{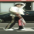}} & 
    \raisebox{-.5\height}{\includegraphics[width=1.0cm]{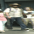}} & 
    \raisebox{-.5\height}{\includegraphics[width=1.0cm]{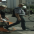}} & 
    
    
    \raisebox{-.5\height}{\includegraphics[width=1.0cm]{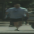}} &
    \raisebox{-.5\height}{\includegraphics[width=1.0cm]{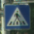}} &
    \raisebox{-.5\height}{\includegraphics[width=1.0cm]{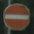}} \\

    Mask &
    \raisebox{-.5\height}{\includegraphics[width=1.0cm]{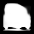}}  &
    \raisebox{-.5\height}{\includegraphics[width=1.0cm]{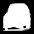}}  &
    \raisebox{-.5\height}{\includegraphics[width=1.0cm]{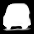}} &
    \raisebox{-.5\height}{\includegraphics[width=1.0cm]{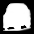}} &
    \raisebox{-.5\height}{\includegraphics[width=1.0cm]{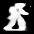}} &
    \raisebox{-.5\height}{\includegraphics[width=1.0cm]{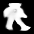}} &
    \raisebox{-.5\height}{\includegraphics[width=1.0cm]{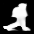}} &
    
    
    \raisebox{-.5\height}{\includegraphics[width=1.0cm]{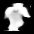}} &
    \raisebox{-.5\height}{\includegraphics[width=1.0cm]{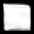}} &
    \raisebox{-.5\height}{\includegraphics[width=1.0cm]{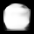}} \\

\end{tabular}
\caption{\textbf{Examples of Cityscapes images and masks generated by our method.} The top row shows the original image, and the middle row is the fake generated by compositing onto a random background with the inferred mask (bottom row). 
}
\label{figure:cityscapes_fake_4_categories}
\end{figure}

Figure \ref{figure:cityscapes_fake_4_categories} shows ``fake images'' created by cut-and-pasting objects using our generated masks.
Generally, the task of generating a realistic composite is well aligned with accurate object segmentation, but there are examples where this is not the case. One such example is the shadows beneath cars, which are important to include in order to synthesize realistic images, but not actually part of the object.


%


\subsection{Effect of Pasting Strategy}
\label{sec:wheretopaste2}

The CityScapes dataset contains object instances at a wide variety of scales corresponding to the wide range of scene depth. For realistic results, it is important to paste objects at the appropriate scale (see Figure \ref{figure:cut_paste_location}). A simple heuristic to achieve this is to paste the object along the same horizontal scanline. We experiment with this approach, shifting with a mean translation of $2\times W$ and standard deviation $W$ (disallowing overlaps), where $W$ is the bounding box width.  This strategy leads to a 4\% absolute increase in per-pixel mask prediction accuracy (from 68\%
to 72\%), when compared to uniformly pasting objects along both the
horizontal and vertical axes. As a sanity check, we also tried pasting Cityscape images into random COCO images for training. This reduced the accuracy to $60\%$ on average and the training process was less stable.

\subsection{Sampling Issues for the Discriminator Network}
\label{sec:sampling_issues}
Convolutional networks are highly sensitive to low-level image statistics, and unintended subtle cues may allow them to ``cheat'', rather than solving the intended problem. An example is described in~\cite{doersch2015unsupervised}, where a convnet used chromatic aberration cues to judge image position. We find a similar effect with sampling artifacts in our approach. In particular, we find that pasting with a mask at lower resolution than the source/destination images leads to a significant drop in performance. In our final implementation we perform compositing at the resolution
of the mask. If we instead attempt to composite at $2 \times$ this resolution, we observe that the performance decreases from 71\% to 66\% in terms of per-pixel mask accuracy. We hypothesize that the discriminator picks up on the additional blurring incurred by the lower resolution mask in real vs fake images in this case.
This suggests that careful image processing is important when dealing with adversarial networks.

\subsection{COCO}




The COCO dataset~\cite{lin2014microsoft} contains a much wider variety of scene content and geometry than our CityScapes and aerial imagery experiments, and the objects typically occupy a much larger fraction of the image. Whilst these appear to be more difficult conditions for our cut+paste approach, we find that our method still works well. 






Since our method requires an object to be pastable within the same image at a new position, we remove objects that are more than $30\%$ of image width as well as very small objects (less than 14 pixels). This results in removing 36\% of the total number of objects, approximately half of which are too small and half too large.
For all instances, we define the ground truth bounding box as the tightest axis-aligned box that encloses the instance mask. We set aside  $15\%$ of the official training set as a development set.


\begin{table}[t!]
\scriptsize
\centering
\begin{tabular}{ | l | C{0.75cm} | C{0.75cm} | C{0.75cm} | C{0.75cm} | C{0.75cm} | C{0.75cm} | C{0.75cm} | C{0.75cm} | C{0.75cm} | C{0.75cm} ||  C{0.75cm} |}
  \hline
        Method & Person & Chair & Car & Cup & Bottle & Book & Bowl & Handbag & Potted plant & Umbrella & {\bf All} \\ \hline
  \multirow{1}{*}{\bf Box}                                          & 0.53      & 0.54      & 0.64      & 0.75      & 0.67      & 0.58      & 0.70      & 0.52      & 0.58      & 0.51      & 0.57 \\ \hline

  

  \multirow{1}{*}{\bf GrabCut \cite{rother2004grabcut}}             & 0.57      & 0.54      & 0.59      & 0.70      & 0.62      & 0.58      & 0.69      & 0.53      & 0.57      & \bb{0.63} & 0.61 \\ \hline

  \multirow{1}{*}{\bf Simple Does It \cite{khoreva2016simple}}      & \bb{0.60} & \bb{0.56} & 0.62      & 0.72      & 0.67      & 0.55      & 0.72      & 0.54      & 0.62      & 0.61      & 0.62  \\ \hline
  \multirow{1}{*}{\bf Cut\&Paste (Ours)}                            & \bb{0.60} & \bb{0.56} & \bb{0.66} & \bb{0.78} & \bb{0.74} & \bb{0.61} & \bb{0.77} & \bb{0.58} & \bb{0.65} & 0.61      & \bb{0.64}\\ \hline \hline
  \multirow{1}{*}{\bf FullySupervised}                              & 0.70      & 0.63      & 0.75      & 0.83      & 0.79      & 0.67      & 0.81      & 0.63      & 0.70      & 0.67      & 0.70\\ \hline
\end{tabular}\vspace{1mm}
\caption{\textbf{mIoU performance on the 10 most common COCO categories.} The final column shows average performance across all 80 categories.}
\vspace{-3mm}
\label{table:coco_results}
\end{table}

\begin{figure}[tb]
\centering
\scriptsize
\centering
\begin{tabular}{ c@{\hspace{0.005\textwidth}}c@{\hspace{0.005\textwidth}}c}
\begin{tabular}{ c@{\hspace{0.005\textwidth}}c@{\hspace{0.005\textwidth}}c@{\hspace{0.005\textwidth}}c }
        Image & GT  & \cite{khoreva2016simple} & Ours \\

        \includegraphics[width=1.025cm]{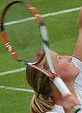} &
        \includegraphics[width=1.025cm]{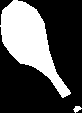} &
        \includegraphics[width=1.025cm]{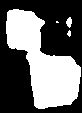} &
        \includegraphics[width=1.025cm]{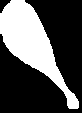} \\ [-0.5ex] 

        \includegraphics[width=1.025cm]{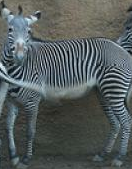} &
        \includegraphics[width=1.025cm]{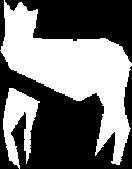} &
        \includegraphics[width=1.025cm]{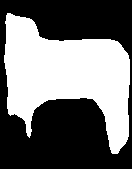} &
        \includegraphics[width=1.025cm]{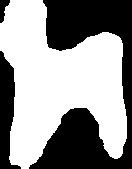} \\ 
\end{tabular} &
\begin{tabular}{ c@{\hspace{0.005\textwidth}}c@{\hspace{0.005\textwidth}}c@{\hspace{0.005\textwidth}}c }
        Image & GT  & \cite{khoreva2016simple} & Ours \\
        \includegraphics[width=0.95cm]{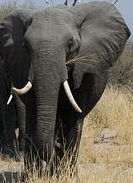} &
        \includegraphics[width=0.95cm]{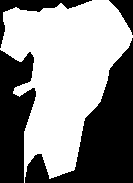} &
        \includegraphics[width=0.95cm]{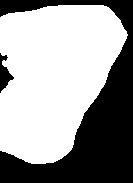} &
        \includegraphics[width=0.95cm]{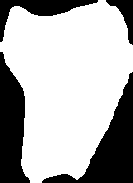} \\ [-0.5ex]
        
        \includegraphics[width=0.95cm]{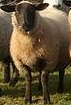} &
        \includegraphics[width=0.95cm]{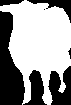} &
        \includegraphics[width=0.95cm]{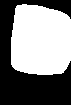} &
        \includegraphics[width=0.95cm]{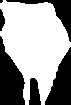} \\ 
\end{tabular} &
\begin{tabular}{ c@{\hspace{0.005\textwidth}}c@{\hspace{0.005\textwidth}}c@{\hspace{0.005\textwidth}}c }
        Image & GT  & \cite{khoreva2016simple} & Ours \\
        \includegraphics[width=0.8cm]{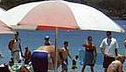} &
        \includegraphics[width=0.8cm]{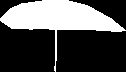} &
        \includegraphics[width=0.8cm]{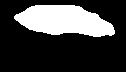} &
        \includegraphics[width=0.8cm]{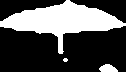} \\ [-0.5ex] 

        \includegraphics[width=0.8cm]{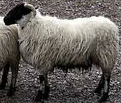} &
        \includegraphics[width=0.8cm]{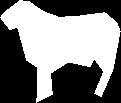} &
        \includegraphics[width=0.8cm]{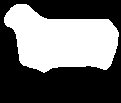} &
        \includegraphics[width=0.8cm]{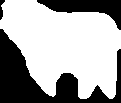} \\ [-0.5ex] 
        
        \includegraphics[width=0.8cm]{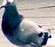} &
        \includegraphics[width=0.8cm]{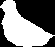} &
        \includegraphics[width=0.8cm]{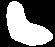} &
        \includegraphics[width=0.8cm]{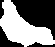} \\ [-0.5ex]

        \includegraphics[width=0.8cm]{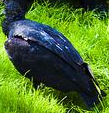} &
        \includegraphics[width=0.8cm]{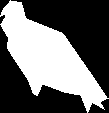} &
        \includegraphics[width=0.8cm]{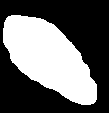} &
        \includegraphics[width=0.8cm]{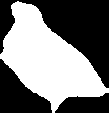} \\ 
\end{tabular} \\
\end{tabular}
\caption{\textbf{COCO examples.} From left to right: the original image, the ground truth mask (GT), the mask predicted by Simple Does It \cite{khoreva2016simple}, and our mask.}
\label{figure:coco_examples}
\end{figure}

\begin{figure}[t]
\centering
\scriptsize
\centering
\begin{tabular}{ c@{\hspace{0.003\textwidth}}c@{\hspace{0.003\textwidth}}c@{\hspace{0.003\textwidth}}c@{\hspace{0.003\textwidth}}c@{\hspace{0.003\textwidth}}c@{\hspace{0.003\textwidth}}c@{\hspace{0.003\textwidth}}c@{\hspace{0.003\textwidth}}c@{\hspace{0.003\textwidth}}c@{\hspace{0.003\textwidth}}c@{\hspace{0.003\textwidth}}c@{\hspace{0.003\textwidth}}c@{\hspace{0.003\textwidth}}c@{\hspace{0.003\textwidth}}c@{\hspace{0.003\textwidth}}c }

    \includegraphics[height=1.6cm]{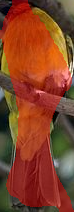} &
    \includegraphics[height=1.6cm]{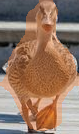} &
    \includegraphics[height=1.6cm]{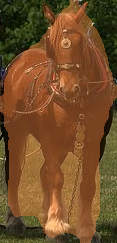} &
    \includegraphics[height=1.6cm]{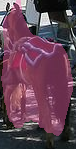} &
    \includegraphics[height=1.6cm]{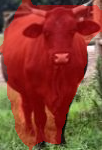} &
    \includegraphics[height=1.6cm]{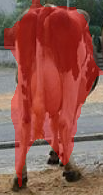} &
    \includegraphics[height=1.6cm]{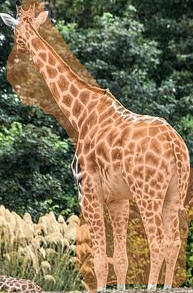} &
    \includegraphics[height=1.6cm]{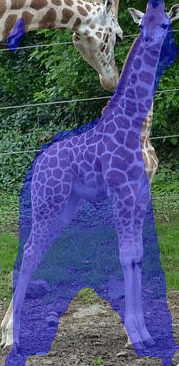} &
    \includegraphics[height=1.6cm]{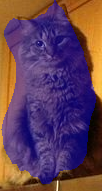} & 
    \includegraphics[height=1.6cm]{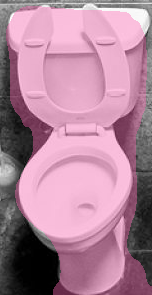} & 
    \includegraphics[height=1.6cm]{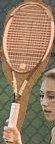} & 
    \includegraphics[height=1.6cm]{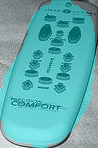} & 
    \includegraphics[height=1.6cm]{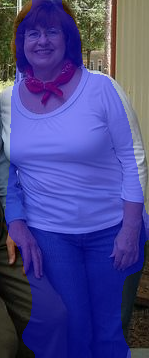} & 
    \includegraphics[height=1.6cm]{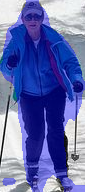} \\ [-0.5ex] 

\end{tabular}    
\begin{tabular}{ c@{\hspace{0.003\textwidth}}c@{\hspace{0.003\textwidth}}c@{\hspace{0.003\textwidth}}c@{\hspace{0.003\textwidth}}c@{\hspace{0.003\textwidth}}c@{\hspace{0.003\textwidth}}c@{\hspace{0.003\textwidth}}c@{\hspace{0.003\textwidth}}c@{\hspace{0.003\textwidth}}c@{\hspace{0.003\textwidth}}c@{\hspace{0.003\textwidth}}c@{\hspace{0.003\textwidth}}c@{\hspace{0.003\textwidth}}c@{\hspace{0.003\textwidth}}c@{\hspace{0.003\textwidth}}c }
    
    \includegraphics[height=1.63cm]{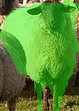} &
    \includegraphics[height=1.63cm]{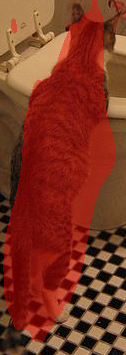} &
    \includegraphics[height=1.63cm]{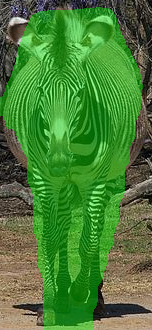} &
    \includegraphics[height=1.63cm]{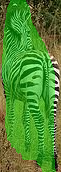} &
    \includegraphics[height=1.63cm]{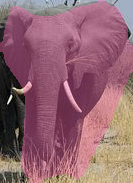} &
    \includegraphics[height=1.63cm]{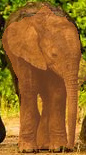} &
    \includegraphics[height=1.63cm]{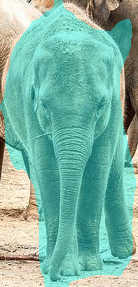} &
    \includegraphics[height=1.63cm]{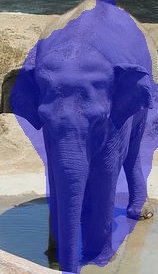} &
    \includegraphics[height=1.63cm]{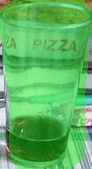} &
    \includegraphics[height=1.63cm]{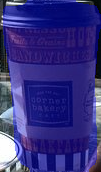} &
    \includegraphics[height=1.63cm]{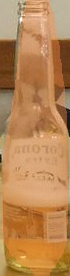} & 
    \includegraphics[height=1.63cm]{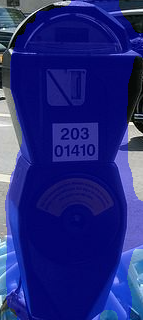} &
    \includegraphics[height=1.63cm]{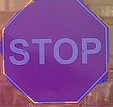} \\ [-0.5ex] 

\end{tabular}
\begin{tabular}{ c@{\hspace{0.003\textwidth}}c@{\hspace{0.003\textwidth}}c@{\hspace{0.003\textwidth}}c@{\hspace{0.003\textwidth}}c@{\hspace{0.003\textwidth}}c@{\hspace{0.003\textwidth}}c@{\hspace{0.003\textwidth}}c@{\hspace{0.003\textwidth}}c@{\hspace{0.003\textwidth}}c@{\hspace{0.003\textwidth}}c@{\hspace{0.003\textwidth}}c@{\hspace{0.003\textwidth}}c@{\hspace{0.003\textwidth}}c@{\hspace{0.003\textwidth}}c@{\hspace{0.003\textwidth}}c@{\hspace{0.003\textwidth}}c@{\hspace{0.003\textwidth}}c@{\hspace{0.003\textwidth}}c@{\hspace{0.003\textwidth}}c@{\hspace{0.003\textwidth}}c@{\hspace{0.003\textwidth}}c@{\hspace{0.003\textwidth}}c@{\hspace{0.003\textwidth}}c@{\hspace{0.003\textwidth}}c@{\hspace{0.003\textwidth}}c@{\hspace{0.003\textwidth}}c@{\hspace{0.003\textwidth}}c@{\hspace{0.003\textwidth}}c@{\hspace{0.003\textwidth}}c@{\hspace{0.003\textwidth}}c }
    
    \includegraphics[height=0.89cm]{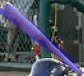} & 
    \includegraphics[height=0.89cm]{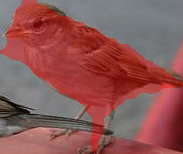} &
    \includegraphics[height=0.89cm]{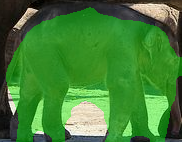} &
    \includegraphics[height=0.89cm]{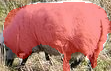} &
    \includegraphics[height=0.89cm]{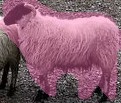} &
    \includegraphics[height=0.89cm]{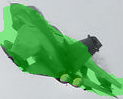} &
    \includegraphics[height=0.89cm]{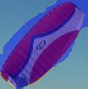} &
    \includegraphics[height=0.89cm]{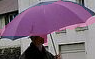} & 
    \includegraphics[height=0.89cm]{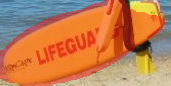} & 
    \includegraphics[height=0.89cm]{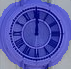} \\ [-0.5ex] 
    
\end{tabular}
\caption{\textbf{COCO examples.} Examples of the masks produced by our method.}
\label{figure:coco_examples_overlay}
\end{figure}

Table \ref{table:coco_results} presents the results for the 10 most common COCO classes, and summary results for all 80 classes. These models were trained using $w_{cls}=w_{cut}=0$.
%
%
Our method exceeds the performance of {\bf GrabCut} in all cases, and {\bf Simple Does It}~\cite{khoreva2016simple} in 70\% of all COCO classes.
We perform particularly well in comparison to [2] on ``baseball bat" (0.43 vs 0.32 mIoU) and ``skis" (0.27 vs 0.23 mIoU). These objects occupy a small fraction of the bounding box, which is problematic for~\cite{khoreva2016simple}, but fine for our method. We perform less well on ``kite" (0.51 vs 0.56 mIoU) and ``airplane" (0.48 vs 0.55). This is perhaps due to the uniform backgrounds that are common for these classes, which will reduce the training signal we see from the cut-and-paste operation (the boundary is arbitrary when pasting with identical backgrounds).
See Figures~\ref{figure:coco_examples} and~\ref{figure:coco_examples_overlay} for examples of masks produced by our method and comparison to those produced by {\bf Simple Does It}.

\subsection{Aerial Imagery}
\label{sec:aerial_results}


To demonstrate the effectiveness of our method in a different setting, we experiment with building segmentation using
a proprietary dataset of aerial images
consisting of $1000\!\times\!1000$ image tiles with annotated building masks. From this dataset, we select a subset of images each of which
contain no
more than 15 houses (in order to allow space in the same image for
pasting), yielding a dataset with 1 million instances.  We also 
similarly generate a validation set containing 
2000 instances. The large size of this dataset also allows us to test performance gains as a function dataset size.   


For these experiments, we trained a Faster R-CNN Inception Resnet v2
(atrous) house detector using the TensorFlow Object Detection API
~\cite{huang2016speed} 
to be used as a backbone for feature extraction.
Since our aerial images are taken at a single scale and
orthorectified, we paste objects into images at locations selected uniformly at random in 
both $x$ and $y$ directions, rejecting pasting locations that 
overlap with other bounding boxes in the image.

\begin{figure}[tb]
\centering
\begin{tabular}{ c c }
    \includegraphics[width=6cm,height=3cm]{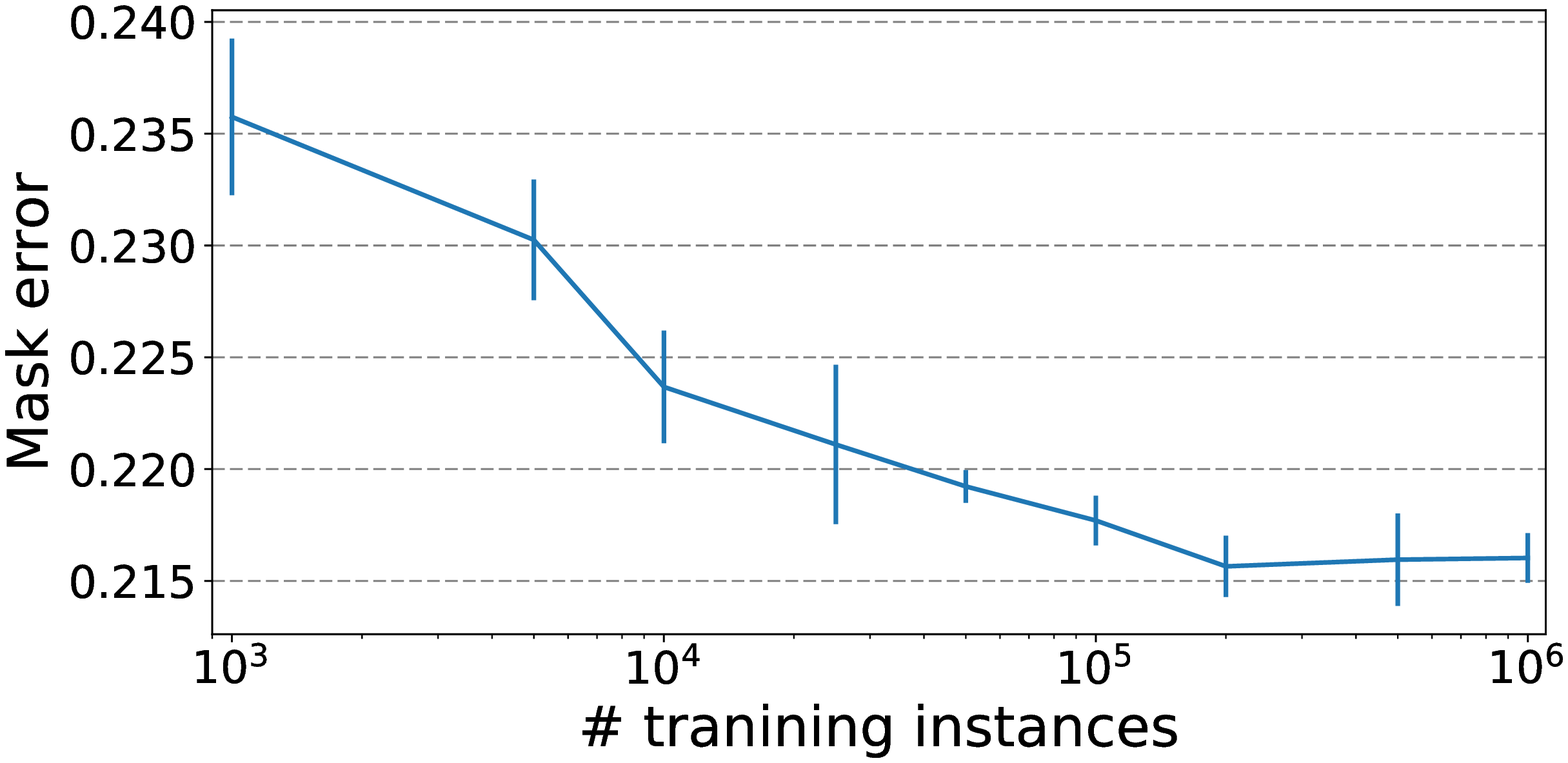} &
    \includegraphics[width=6cm,height=3cm]{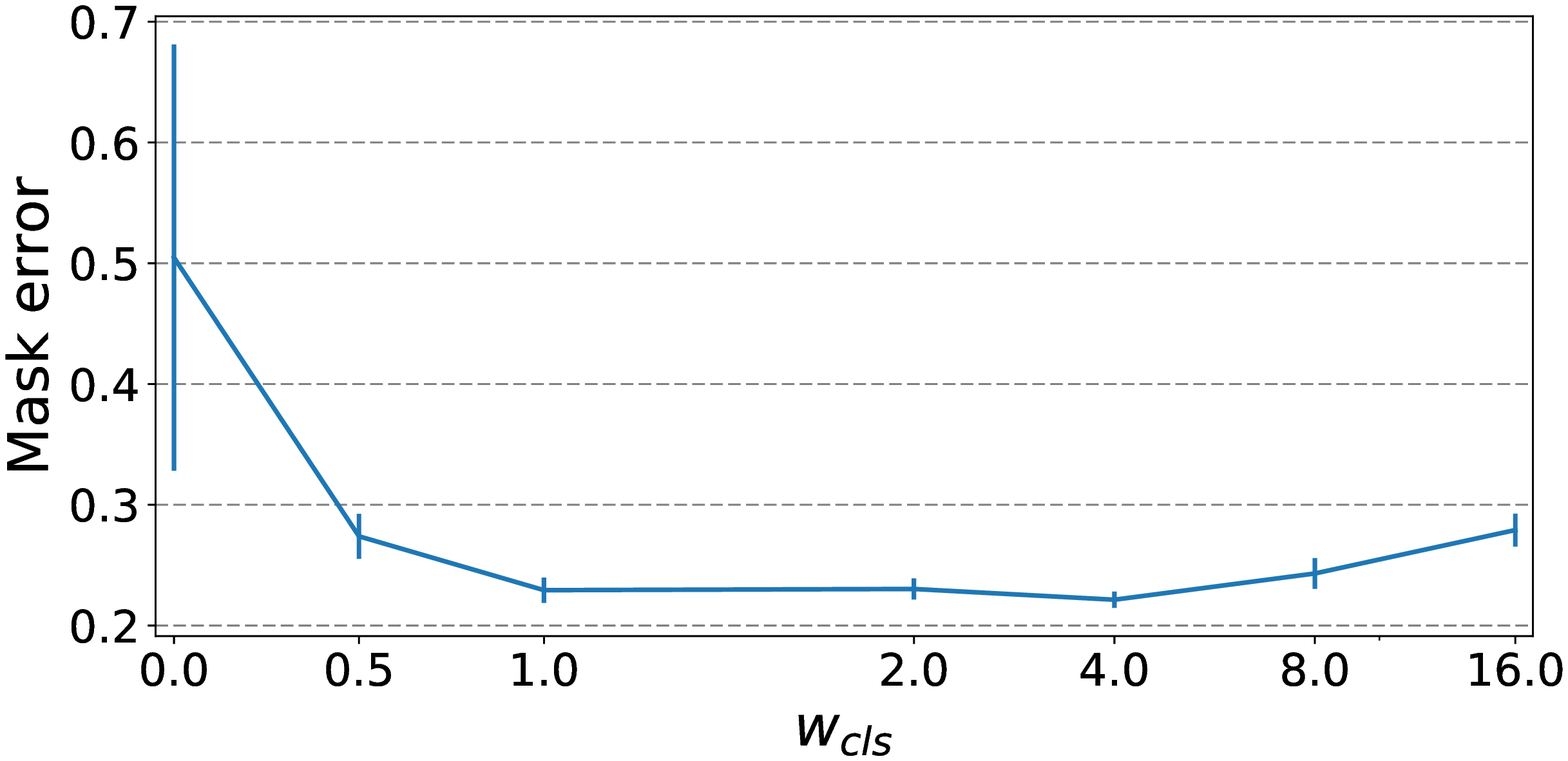} \\
  
  (a) & (b) \\
  \end{tabular}
  \caption{
  \textbf{Effect of classification loss and dataset scale on mask accuracy.} 
  (a) demonstrates the effect of scale of the training set
  on mask accuracy for aerial imagery.
   (b) demonstrates the effect of classification loss weight $w_{cls}$ on the accuracy of the predicted masks; curves show mean and standard deviation over 4 runs for (a) and 5 models for (b).}
  \label{figure:aerial_plots}
\end{figure}

\paragraph{Effect of Dataset Scale.} 
Figure \ref{figure:aerial_plots}(a) shows the effect of data size on the average performance of our models. 
Increasing data size helps the training process, increasing the number of training instances from 5K to 1M reduces the mask prediction error by about $10\%$. 

\begin{figure}[tb]
\centering
\begin{tabular}{ c c c c }

        \includegraphics[width=0.24\textwidth]{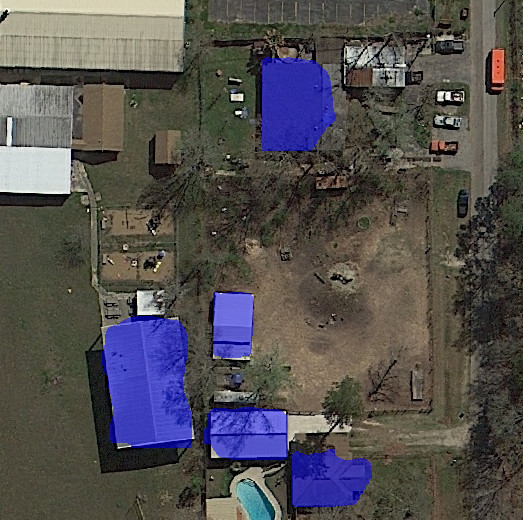} &
        \includegraphics[width=0.24\textwidth]{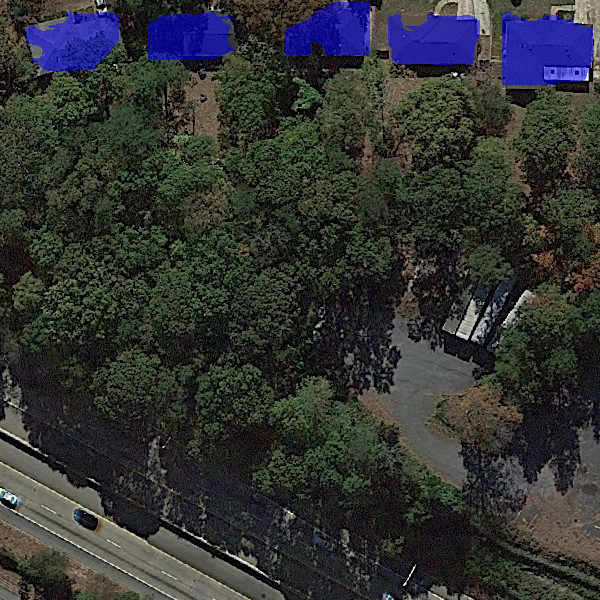} &
        \includegraphics[width=0.24\textwidth]{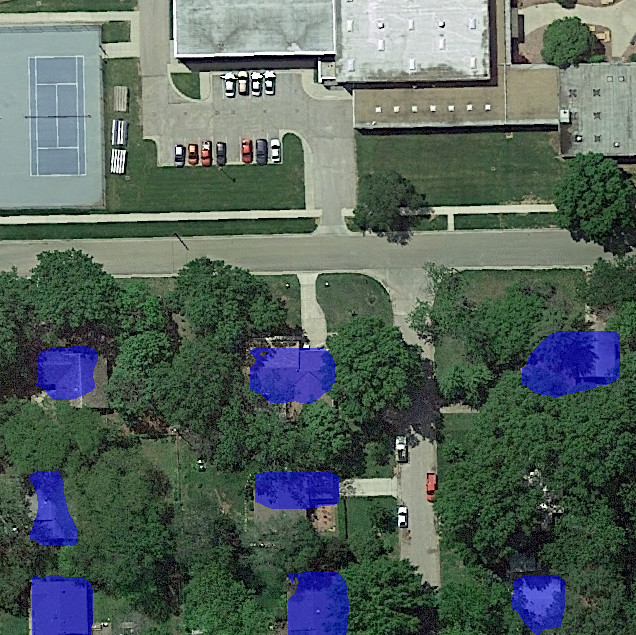} &
        \includegraphics[width=0.24\textwidth]{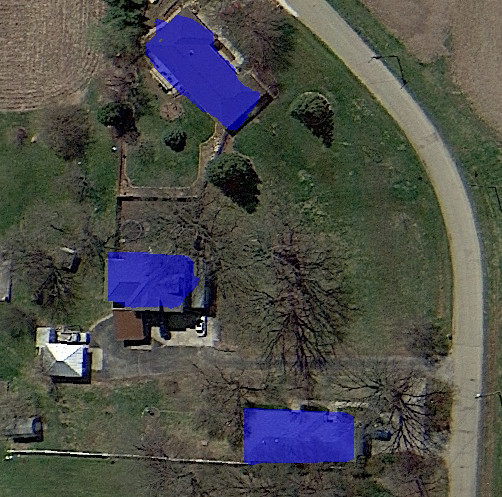} \\
        
        
\end{tabular}
\caption{\textbf{Aerial imagery examples.} Examples of masks produced by our method.}
\label{figure:aerial_overlays}
\end{figure}

\paragraph{Effect of Loss Weightings.} Figure \ref{figure:aerial_plots}(b) shows the effect of the classification loss weight $w_{cls}$ on the overall performance of the model. 
With no classification loss ($w_{cls}=0$) the performance is poor and the model is unstable, as indicated by the error bars. With increasing classification loss, performance improves and the error bars become tighter showing the training process is much more stable. The optimal weight in this case is in the range of $w_{cls}\in[1,4]$.  When conducting a similar experiment for the Cityscapes dataset we found that the classification weight increases stability but does not improve performance overall.
This may be due to the high incidence of occlusion in our aerial image data, e.g., a subsection of a mask often resembles a building occluded by trees. Figure \ref{figure:aerial_overlays} shows a few examples of typical aerial images and the segmentation masks our method produces when trained using $w_{cls}=1$. 

We also experimented with various weightings of cut loss $w_{cut}$, but found this did not work as well as our classification loss.
We note that while cut loss serves to expand masks that are too small, once the mask is larger than the object the cut loss provides no useful signal. 


\subsection{Failure Cases}

\begin{figure}[tb]
\centering
\footnotesize
\begin{tabular}{ r@{\hspace{0.005\textwidth}}c@{\hspace{0.005\textwidth}}c@{\hspace{0.005\textwidth}}c@{\hspace{0.005\textwidth}}c@{\hspace{0.005\textwidth}}c@{\hspace{0.015\textwidth}}c@{\hspace{0.005\textwidth}}c@{\hspace{0.005\textwidth}}c@{\hspace{0.005\textwidth}}c@{\hspace{0.005\textwidth}}c }

    \vspace{0.5mm}Real &
    \raisebox{-.5\height}{\includegraphics[width=1.0cm]{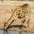}} &
    \raisebox{-.5\height}{\includegraphics[width=1.0cm]{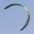}} &
    \raisebox{-.5\height}{\includegraphics[width=1.0cm]{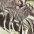}} &
    \raisebox{-.5\height}{\includegraphics[width=1.0cm]{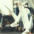}} &
    \raisebox{-.5\height}{\includegraphics[width=1.0cm]{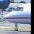}} &
    
    \raisebox{-.5\height}{\includegraphics[width=1.0cm]{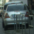}} &
    \raisebox{-.5\height}{\includegraphics[width=1.0cm]{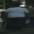}} &
    \raisebox{-.5\height}{\includegraphics[width=1.0cm]{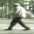}} &
    \raisebox{-.5\height}{\includegraphics[width=1.0cm]{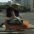}} &
    \raisebox{-.5\height}{\includegraphics[width=1.0cm]{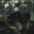}} \\
    
    \vspace{0.5mm}Fake &
    \raisebox{-.5\height}{\includegraphics[width=1.0cm]{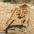}} &
    \raisebox{-.5\height}{\includegraphics[width=1.0cm]{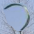}} &
    \raisebox{-.5\height}{\includegraphics[width=1.0cm]{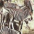}} &
    \raisebox{-.5\height}{\includegraphics[width=1.0cm]{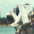}} &
    \raisebox{-.5\height}{\includegraphics[width=1.0cm]{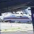}} &
    
    \raisebox{-.5\height}{\includegraphics[width=1.0cm]{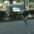}} &
    \raisebox{-.5\height}{\includegraphics[width=1.0cm]{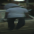}} &
    \raisebox{-.5\height}{\includegraphics[width=1.0cm]{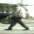}} &
    \raisebox{-.5\height}{\includegraphics[width=1.0cm]{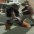}} &
    \raisebox{-.5\height}{\includegraphics[width=1.0cm]{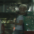}} \\
    
    Our mask &
    \raisebox{-.5\height}{\includegraphics[width=1.0cm]{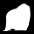}} &
    \raisebox{-.5\height}{\includegraphics[width=1.0cm]{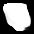}} &
    \raisebox{-.5\height}{\includegraphics[width=1.0cm]{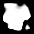}} &
    \raisebox{-.5\height}{\includegraphics[width=1.0cm]{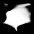}} &
    \raisebox{-.5\height}{\includegraphics[width=1.0cm]{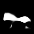}} &
    
    \raisebox{-.5\height}{\includegraphics[width=1.0cm]{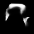}} &
    \raisebox{-.5\height}{\includegraphics[width=1.0cm]{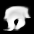}} &
    \raisebox{-.5\height}{\includegraphics[width=1.0cm]{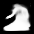}} &
    \raisebox{-.5\height}{\includegraphics[width=1.0cm]{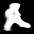}} &
    \raisebox{-.5\height}{\includegraphics[width=1.0cm]{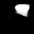}} \\

\end{tabular}
\caption{\textbf{Failures.} COCO images are on the left and Cityscapes are on the right.}
\label{figure:coco_failures}
\end{figure}

A few failure cases of our method are presented in Figure \ref{figure:coco_failures}. For the Giraffe and Kite examples, the mask is overestimated, but the non-unique backgrounds lead to convincing fake images. Note that the shadow of the Giraffe is copied in the first case, another common failure mode.
Other examples include missing or added parts, such as the missing head and extra ``skirt" connecting the legs for the people towards the right of the figure. Next to these is an interesting example with a person carrying a suitcase. The generator decides it would be more realistic to change this to legs in this example, presumably because suitcases are a rare occurrence in the training set. 
Note that the ``fake" images are often still realistic in many of these cases.

\section{Conclusions}
We have presented a new approach to instance segmentation that uses a simple property of objects, namely that they are ``cut-and-pastable", coupled with a generative adversarial network to learn object masks. Our method exceeds the performance of existing box-supervised methods on the CityScapes and COCO datasets, with no mask ground truth and without the need for pre-trained segment or boundary detectors.


We have shown that intelligent object placement in the paste step can significantly improve mask estimation. This suggests an interesting direction for future work, where the compositing step is also data-dependent. For example, object placement, colour and illumination could depend on the destination image. Related work shows this works well for data augmentation~\cite{dwibedi2017cut,georgakis2017synthesizing,alhaija2017augmented}. 

More generally, and as in work such as~\cite{tung2017adversarial}, we could envisage a range of settings where vision+graphics imitate photography, with adversarial losses to jointly optimise image understanding and rendering stages. Such methods could open up the possibility of performing detailed visual perception with reduced dependence on large-scale supervised datasets.

\vspace{4mm}

\noindent\textbf{Acknowledgments:} 
We thank our colleagues David Lowe,  Kevin Murphy, \mbox{Rodrigo Benenson}, George Papandreou,  Alireza Fathi, Tom Weng, Robert Gens, and Andrey Zhmoginov for useful discussions and suggestions on the manuscript.

\clearpage

\bibliographystyle{splncs}
\bibliography{egbib}
\end{document}